# NorBERTo: A ModernBERT Model Trained for Portuguese with 331 Billion Tokens Corpus

**Enzo S. N. Silva**[†1], **Pablo B. Costa**[1], **Raphael C. Vlasman**[1], **Rosimeire P. Costa**[2], **Henrique L. P. Silva**[2], **Lucas F. A. O. Pellicer**[2], **Guilherme Rinaldo**[2], **Renato A. Almeida**[2], **Darian S. R. Rabbani**[2], **Cinthya O. Oestreich**[2], **Vinicius F. Caridá**[1]

[1]Itaú Unibanco, São Paulo, SP, Brasil
[2]ICTi, São Paulo, SP, Brasil

{enzo.silva22, pablo.costa, raphael.vlasman, rosimeire.pereira-costa, henrique.a.pereira-silva, lucas.pellicer, guilherme.rinaldo, renato-augusto.almeida, darian.rabbani, cinthya.oestreich-silva, vinicius.carida}@itau-unibanco.com.br

## Abstract

High-quality corpora are essential for advancing Natural Language Processing (NLP) in Portuguese. Building on previous encoder-only models such as BERTimbau and Albertina PT-BR, we introduce NorBERTo, a modern encoder based on the ModernBERT architecture, featuring long-context support and efficient attention mechanisms. NorBERTo is trained on Aurora-PT, a newly curated Brazilian Portuguese corpus comprising 331 billion GPT-2 tokens collected from diverse web sources and existing multilingual datasets. We systematically benchmark NorBERTo against strong baselines on semantic similarity, textual entailment and classification tasks using standardized datasets such as ASSIN 2 and PLUE. On PLUE, NorBERTo-large achieves the best results among the encoder models we evaluated, notably reaching 0.9191 F1 on MRPC and 0.7689 accuracy on RTE. On ASSIN 2, NorBERTo-large attains the highest entailment F1 (~0.904) among all encoders considered, although Albertina-900M and BERTimbau-large still hold an advantage. To the best of our knowledge, Aurora-PT is currently the largest openly available monolingual Portuguese corpus, surpassing previous resources. NorBERTo provides a modern, mid-sized encoder designed for realistic deployment scenarios: it is straightforward to fine-tune, efficient to serve, and well suited as a backbone for retrieval-augmented generation and other downstream Portuguese NLP systems.

## 1 Introduction

Pre-trained transformer encoders such as BERT (Devlin et al., 2018) have transformed NLP by providing contextual representations that can be efficiently adapted to a wide range of tasks, from natural language inference and question answering to sentiment classification. Bidirectional masked-language pre-training has proved particularly effective at capturing sentence-level semantics, yielding large gains over non-contextual or feature-based approaches. In Portuguese, BERT-based models such as BERTimbau (Souza et al., 2020) have delivered substantial gains in Named Entity Recognition (NER), sentiment analysis, and textual entailment on benchmarks including ASSIN 2 and related tasks (Finardi et al., 2021; Warner et al., 2024).

In parallel, Large Language Models (LLMs) have popularized the idea of solving many tasks via prompting and in-context learning. These models excel at open-ended text generation and cross-domain generalization, but they also pose significant challenges: high computational and monetary cost, latency constraints, privacy and governance issues for sensitive data, and the well-documented problem of hallucinations and unreliable reasoning (Kostikova et al., 2025). Empirical studies further show that, in many structured or domain-specific settings, smaller, task-specific models can match or even surpass general-purpose LLMs, especially when accuracy-to-cost, calibration, or fairness are taken into account (Souza et al., 2020; Sanh et al., 2019).

At the same time, encoder-only models remain highly competitive for discriminative tasks such as classification, retrieval and ranking. Recent work such as ModernBERT (Warner et al., 2024) shows that updating BERT's architecture with Rotary Positional Embeddings (RoPE), long-context attention, and optimized implementations (e.g., FlashAt-tention, sequence packing) can yield substantial gains in efficiency and downstream accuracy, even compared to larger or older encoders.

For Portuguese, however, encoder development still lags behind English in two key respects: (i) the availability of very large, curated monolingual corpora, and (ii) the adoption of modern encoder architectures with long-context support and efficient attention. BERTimbau, for instance, is trained on approximately 2.68 billion tokens from Brazilian Web as Corpus (brWaC) (Wagner Filho et al., 2018) and Wikipedia, while Albertina (Rodrigues et al., 2023) leverages somewhat larger corpora but remains far below the scale now common for English. Recent work on the decoder-only Tucano models (Corrêa et al., 2024, 2025) shows that scaling Portuguese pre-training to approximately 200 billion tokens, using the GigaVerbo corpus, can substantially improve generative models, suggesting that similar gains may be attainable for encoders.

This work offers three main contributions to Portuguese NLP: (i) the creation and publication of a new monolingual corpus in Brazilian Portuguese containing 331 billion tokens named Aurora-PT[1], carefully cleaned, deduplicated, and filtered to remove toxic and low-quality content; (ii) the training of a new BERT-type model called NorBERTo[2], publicly available, based on the optimized ModernBERT architecture (Warner et al., 2024), using our corpus Aurora-PT; and (iii) a complete pipeline for dataset creation and model training, including the comparison of NorBERTo's performance with existing state-of-the-art Portuguese models (e.g., BERTimbau and Albertina PT-BR) on standardized benchmarks such as PLUE (Portuguese Language Understanding Evaluation) and ASSIN 2 (Semantic Similarity and Textual Inference). The goal is to evaluate how large-scale data and architectural advances impact performance, and whether a compact encoder can match or surpass larger models or LLMs in Portuguese language understanding.

This article adopts the scientific method. The Introduction contextualizes advances in Portuguese language models and emphasizes the importance of high-quality data and modern architectures. Section 2 reviews BERT and its Portuguese successors (BERTimbau, BERTaú, Albertina), recent large language models, and relevant corpora and evaluations, also discussing the limitations of LLMs and the advantages of smaller models for specific tasks. Section 3 details the construction of the Aurora-PT corpus and the training of the NorBERTo model, highlighting our distinctive methodology. Section 4 presents preliminary results, comparing NorBERTo to baseline models on key benchmarks, supported by tables and figures. Finally, Section 5 offers final considerations and future perspectives.

[1] Available at: Hugging Face – Aurora-PT

[2] Available at: Hugging Face – NorBERTo

## 2 Related Work

In this section, we present a review of related work. We examine studies focused on Portuguese language models and the linguistic resources employed for their training.

### 2.1 Encoder-based Pre-trained Language Models for Portuguese

BERT marked a paradigm shift in NLP by leveraging the transformer encoder architecture (Devlin et al., 2018). BERT employs a deep, bidirectional transformer encoder, allowing it to capture context from both left and right of each token simultaneously. BERT's pretraining strategy, based on Masked Language Modeling (MLM) and Next Sentence Prediction (NSP), allows it to develop a nuanced understanding of language structure, making it highly effective for NLP tasks (Devlin et al., 2018).

BERT's success led to numerous adaptations and specialized models for various languages, including Portuguese. Early models such as BERTimbau were trained on the brWaC corpus, a massive collection of Brazilian web pages totaling approximately 2.68 billion tokens (Souza et al., 2020). This monolingual approach enabled BERTimbau to outperform multilingual models like BERT in Portuguese-specific tasks, establishing a strong baseline for subsequent research (Souza et al., 2020).

Recent advances have introduced models built on even larger and cleaner datasets. Albertina-PT, for instance, utilizes the DeBERTa encoder architecture and is pretrained on corpora covering both European and Brazilian Portuguese, including brWaC and additional open datasets (Rodrigues et al., 2023). PeLLE models are based on the RoBERTa architecture and trained exclusively on the Corpus Carolina, a carefully curated and openly licensed dataset of Brazilian Portuguese texts (Mello et al., 2024). These initiatives highlight the importance of both data quality and legal permissiveness in model development.

A domain-specific variant such as BERTaú stands out as a recent encoder model for Portuguese. It was trained from scratch with data from the Itaú virtual assistant chatbot solution. BERTaú leverages modern architectural choices and a tokenizer tailored to the language, further advancing the performance of monolingual models (Finardi et al., 2021). Collectively, these models demonstrate that leveraging large, high-quality, and language-specific corpora is key to achieving state-of-the-art results in Portuguese, with monolingual models consistently outperforming their multilingual counterparts in targeted benchmarks.

In comparison to the aforementioned BERT models, ModernBERT represents a leap forward in architectural innovation and data scale. It incorporates features such as RoPE, long-context support (up to 8192 tokens), and efficient attention mechanisms (Warner et al., 2024), and comprises tens of billions of tokens from web crawls and parallel data. This enables the model to process entire documents and technical texts, including code, without truncation. Although ModernBERT is multilingual, it enables the development of stronger Portuguese NLP models.

### 2.2 Challenges and Limitations of Generative Language Models

Beyond encoder-only models such as BERT, the transformer architecture (Vaswani et al., 2017) enabled the development of decoder-based LLMs. A key advantage of LLMs lies in their ability to address diverse tasks without requiring task-specific training. By modifying the initial prompt, an LLM can perform summarization, translation, or text classification (Qin et al., 2025). This flexibility positions LLMs as powerful tools for a wide range of NLP applications.

However, several strands of work highlight important limitations and trade-offs. Due to the scale of LLMs, deployment cost and model latency can be highly significant (Kostikova et al., 2025). Recent surveys have highlighted the risk of hallucinations and the generation of incorrect data (Ji et al., 2023). Furthermore, general-purpose LLMs tend to be more limited in specific domains, such as clinical prediction, automated scoring, and tabular classification, when compared to specialized, smaller-scale models (Kostikova et al., 2025).

These findings support the idea that LLMs and smaller models are complementary. LLMs excel at open-ended generation and cross-domain reasoning, while compact encoders and domain-specific models remain preferable for many focused, high-throughput, or safety-critical tasks. Our work follows this “right-sized AI” perspective by introducing a mid-sized encoder tailored to Portuguese, rather than another general-purpose LLM.

### 2.3 Corpora for Training Portuguese Language Models

The availability of large-scale text corpora in Portuguese is essential for training robust language models. Historically, the brWaC (Wagner Filho et al., 2018) marked a significant milestone as one of the first extensive collections for Brazilian Portuguese. It contained approximately 2.68 billion tokens extracted from web content (~ 25 GB of text). Another notable resource is the Carolina Corpus (Crespo et al., 2023), which focuses on informal language and internet slang in European Portuguese, with approximately 0.82 billion tokens. While these datasets were important milestones, they were much smaller than English or Chinese datasets, limiting Portuguese model competitiveness.

Recently, efforts to expand Portuguese corpora have accelerated. The Tucano Corpus (Corrêa et al., 2024, 2025) introduced innovative curation methods for cleaner, more diverse data, while the FineWeb Datasets (Penedo et al., 2025) applied advanced pipelines to produce refined corpora with up to 50 billion tokens. These initiatives highlight the growing importance of sophisticated data processing for building scalable and effective resources.

Researchers also introduced the Aroeira Corpus (Lira et al., 2025), described as "a curated corpus for the Portuguese language with a large number of tokens." Built using texts from various internet platforms (primarily Common Crawl), Aroeira underwent a rigorous cleaning and content safety pipeline to ensure high-quality data. The result was a corpus with 15 billion tokens (~100 GB of pure text).

To better illustrate the evolution of Portuguese corpora over time, Table 1 provides a comparative overview of key datasets, including brWaC, Carolina, Aroeira, and our newly developed corpus. This table highlights the exponential growth in size and scope, culminating in our current work. Our latest contribution represents a significant leap forward in this domain. This corpus serves as the foundation for training the NorBERTo model, which aims to push the boundaries of NLP in Portuguese.

| Corpus | Documents Size | # Quantity of Documents |
|---|---|---|
| GigaVerbo-Text-Filter | 0.86 GB | 0.11 mi |
| Wikipédia | 1 GB | 1.1 mi |
| Carolina | 11 GB | 2.11 mi |
| brWaC | 25 GB | 3.53 mi |
| Aroeira | 100 GB | 24.9 mi |
| FineWeb-2_Latn-por | 257 GB | 200 mi |
| GigaVerbo | 780 GB | 145 mi |
| Aurora-PT | 873 GB | 275 mi |

Table 1: Comparative Overview of Portuguese Corpora. brWaC (Wagner Filho et al., 2018), Carolina (Crespo et al., 2023), GigaVerbo-Text-Filter (Corrêa et al., 2024, 2025), Aroeira (Lira et al., 2025), GigaVerbo (Corrêa et al., 2024, 2025), FineWeb-2_Latn-por (Penedo et al., 2025)

## 3 Development of New Portuguese Resources

In this section, we present the two new resources generated by this work: the 331-billion-token Aurora-PT corpus and the NorBERTo model based on the ModernBERT architecture. We discuss details of the development of these two outcomes of our research.

### 3.1 Aurora-PT: 331 Billion Token Portuguese Corpus

As an initial step in training a Portuguese language model, we developed a robust monolingual corpus named Aurora-PT. The construction process was inspired by the pipeline used in FineWeb v1 (Penedo et al., 2024), with a key distinction: instead of relying on Common Crawl dumps, we aggregated nine Portuguese or multilingual datasets annotated by language, as detailed in Table 2.

The preprocessing pipeline comprised several stages. First, we applied language filtering using GlotLID with a threshold of 0.799 for Portuguese (Kargaran et al., 2023). Next, we performed deduplication via MinHash, configured with 112 hashes distributed across 14 buckets. Following deduplication, we adopted the quality filters introduced by FineWeb, which include both C4-based heuristics: the removal of lines with fewer than three words; the removal of lines containing curly brackets; and finally, the removal of lines containing any of the terms: “Javascript”, “cookies” and “lorem ipsum”.

Then FineWeb-specific rules were applied: the removal of documents with fewer than 12% of lines ending in punctuation; the removal of documents where more than 67% of lines contain fewer than 30 characters; and the removal of documents with over 10% of characters in duplicated lines. All processing steps were implemented using the DataTrove library from Hugging Face.

After filtering, the resulting corpus contained approximately 330 billion tokens when tokenized with GPT-2 and about 226 billion tokens using a tokenizer trained on the corpus itself. Using GPT-2 tokenization as a common reference, Aurora-PT is currently the largest monolingual Portuguese dataset.

| Dataset | # Original Documents | # Final Documents | Retention (%) |
|---|---|---|---|
| CC100 | 339889917 | 3919914 | 1.15 |
| mOSCAR | 8033406 | 4725016 | 58.82 |
| Aya | 9247 | 958 | 10.36 |
| Fineweb v2 | 189883678 | 172467090 | 90.83 |
| Blogset-br | 4349657 | 613403 | 14.10 |
| Aroeira | 34841241 | 7275814 | 20.88 |
| mC4 | 169408501 | 79110740 | 46.70 |
| Wikipedia | 1112246 | 391647 | 35.21 |
| HPLT 2.0 | 14581809 | 7107555 | 48.74 |

Table 2: Document retention after filtering for Portuguese datasets.

### 3.2 NorBERTo: Portuguese Modern BERT Trained Model

NorBERTo is a ModernBERT-style (Warner et al., 2024) encoder tailored to Portuguese. It incorporates RoPE for better extrapolation to longer contexts. The model alternates between local and global attention: the first layer and every third layer use global attention, while the others apply local attention within a fixed window, optimizing computational efficiency for long sequences. Additional enhancements include sequence packing and unpadding during training, which prevents computation on padding tokens and improves throughput, as well as gated feed-forward layers (such as GeGLU) and bias-free linear projections.

We train two configurations: NorBERTo-base, with approximately 150M parameters, 22 layers, hidden size 768, 12 attention heads, and a GLU expansion size of 2,304; and NorBERTo-large, with approximately 395M parameters, 28 layers, hidden size 1,024, 16 attention heads, and a GLU expansion size of 5,248. Both variants use the same 8,192-token context and share a 50,368-token vocabulary learned from Aurora-PT. The architectural metrics described above — including the ModernBERT framework, the use of RoPE, the lo-

cal–global attention pattern, and the configurations of the base and large variants — are summarized in Table 3.

The pre-training objective was MLM with a masking rate of 30%, implemented using the Composer library developed by Answer.AI. The training corpus consisted of Aurora-PT, tokenized with NorBERTo tokenizer, totaling 226B tokens.

The training pipeline comprised three phases: The first being the standard pre-training, with 85% of tokens, using a context window of 1,024 tokens; followed by the context extension, using 12.5% of tokens, expanding the context window to 8,192 tokens; then the learning rate decay phase, using 2.5% of tokens, maintaining the extended context window.

To enable context extension, and given the relative scarcity of long documents in Portuguese corpora, documents longer than 1,024 tokens were reserved for phases two and three (Context Extension and Learning Rate Decay, respectively). Additional examples required to meet token quotas were randomly sampled from the remaining documents, and the split between phases two and three was also randomized. Training phase sizes were scaled to match the available data while preserving their relative ratios.

Training was conducted over a single epoch, lasting approximately three days on 8 NVIDIA H100 GPUs with FlashAttention v2 support. Hyperparameters for all phases and models are summarized in Table 4.

| | BASE | LARGE |
|---|---|---|
| Vocab | 50368 | 50368 |
| Camadas | 22 | 28 |
| Hidden Size | 768 | 1024 |
| Attentions Heads | 12 | 16 |
| Global attn | 1º and every third | 1º and every third |
| Local Attn Window | 128 | 128 |
| Intermediate Size | 1152 | 2624 |
| GLU expansion | 2304 | 5248 |
| Rope theta | 160000 | 160000 |
| Local attn rope theta | 10000 | 10000 |

Table 3: Architeture parameters of the trained models

# 4 Experiments and Results

## 4.1 Comparative Analysis of Portuguese Language Corpora

As part of this evaluation, we conducted a comparative analysis of some Portuguese language corpora. For the metrics described below, we used a random sample of 1% from each corpus, except for Gigaverbo-Text-Filter and brWaC, which were analyzed in full due to their comparatively smaller size relative to the remaining corpora.

To assess lexical diversity across these corpora, we employed two widely used metrics: Type-Token Ratio (TTR) and Hypergeometric Distribution D (HD-D). The TTR computes the ratio between types and tokens; however, it is highly sensitive to text length, which can distort comparisons among corpora of different sizes (Templin, 1957; Richards, 1987). To mitigate this effect, we adopted HD-D, a probabilistic sampling-based measure that provides more stable and comparable estimates of lexical diversity, even for corpora with varying lengths (McCarthy and Jarvis, 2010).

When observing the TTR boxplots for the corpora (Figure 1), moderate differences between the medians can be seen, ranging from 0.555 to 0.615. The corpus Aurora-PT shows a median of 0.567, placing it in the intermediate range of the observed values. Its distribution is relatively compact, suggesting stability in lexical diversity across the samples. In the case of the HD-D metric, a similar pattern is observed: the corpus Aurora-PT maintains values aligned with the central group of corpora, indicating that its lexical diversity is consistent and not artificially inflated or reduced by size.

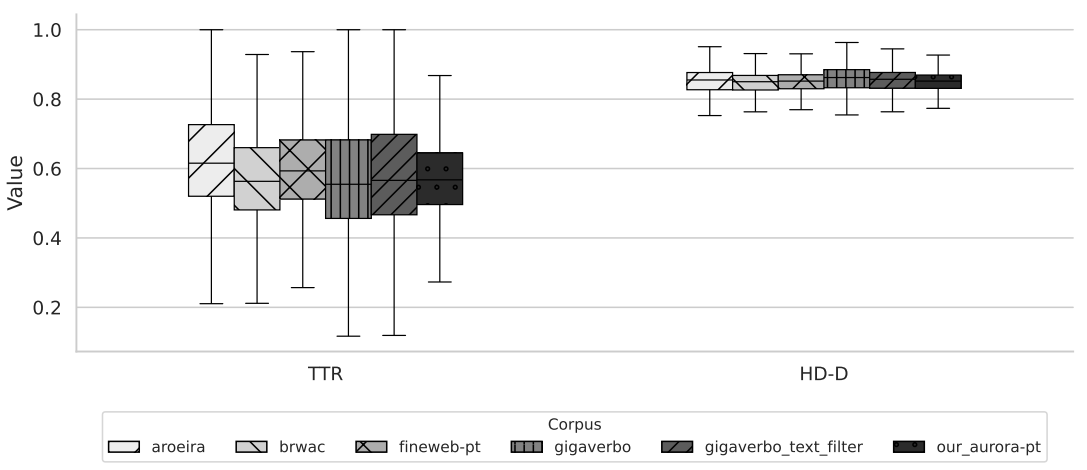

Figure 1: Distribution of TTR and HD-D by corpus.

The metrics Lexical Sophistication and Average Word Frequency are commonly used to assess qualitative aspects of the vocabulary employed in a corpus. Lexical Sophistication measures the proportion of less frequent words, indicating the degree of lexical complexity in the text (Laufer and Nation, 1995; Nation and Nation, 2001). Meanwhile, the Average Word Frequency metric calculates the mean frequency of words in a corpus based on reference lexical databases: higher values reflect greater use of frequent, everyday vocabulary (Brysbaert and New, 2009).

The analysis of the Lexical Sophistication boxplots (Figure 2) reveals relatively stable patterns among the corpora, with moderate variation in the distributions. The corpus Aurora-PT is positioned near the center of the set, showing behavior similar to adjacent corpora. This pattern indicates that the corpus Aurora-PT employs a balanced proportion of rare vocabulary, consistent with the range observed in the other corpora. For the Average Word Frequency metric, a similar behavior is observed: the corpus Aurora-PT remains aligned with the intermediate corpora, characterizing the predominant use of medium-frequency vocabulary.

| | Pre-training | | Context Extension | | LR Decay | |
|---|---|---|---|---|---|---|
| | Base | Large | Base | Large | Base | Large |
| Tokens | 195B | | 29B | | 6B | |
| Max. Context Size | 1024 | | 8192 | | 8192 | |
| Batch Size | 4096 | 4928 | 576 | 616 | 576 | 624 |
| Warmup Tokens | 5B | 5B | – | – | – | – |
| Learning Rate | 8e-4 | 5e-4 | 3e-4 | 5e-5 | 3e-4 | 5e-5 |
| Schedule | trapezoidal | trapezoidal | const. | const. | 1-sqrt | 1-sqrt |
| Warmup | 3B | 3B | – | – | – | – |
| Decay | – | – | – | – | 6B | 6B |
| Initialization | Megatron | Tiling w/ Base | – | – | – | – |
| Dropout (Attention) | 0.1 | | | | | |
| Dropout (Rest) | 0.0 | | | | | |
| Optimizer | StableAdamW | | | | | |
| $\beta_1$ | 0.90 | | | | | |
| $\beta_2$ | 0.98 | | | | | |
| $\epsilon$ | 1e-6 | | | | | |

Table 4: Training hyperparameters for Base and Large models across different phases.

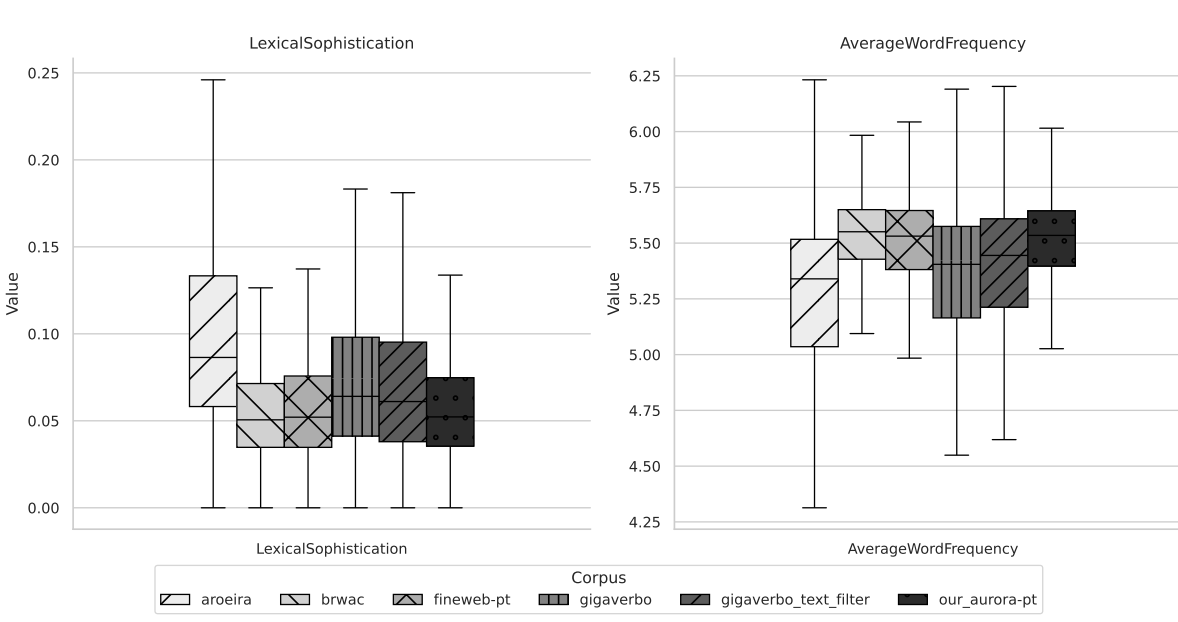

Figure 2: Distribution of Lexical Sophistication and Average Word Frequency by corpus.

An analysis of the Lexical Density (Ure, 1971) and Stopword Ratio (Schütze et al., 2008) metrics was also conducted, both of which describe the lexical composition of the corpora (Table 5). Lexical Density indicates the proportion of content words, whereas Stopword Ratio reflects the proportion of function words.

When analyzing Table 5, it is observed that in the corpus Aurora-PT, both metrics present values very close to those of the other corpora, indicating that its lexical composition follows the same general pattern. The differences between the medians are small, and the interquartile ranges reveal strong consistency, suggesting that there are no significant variations in the proportion of content words and function words.

| **Corpus** | **Lexical Density** | **Stopword Ratio** |
|---|---|---|
| brWaC | 0.464 [0.420- 0.501] | 0.358 [0.332-0.381] |
| GigaVerbo | 0.447 [0.387-0.496] | 0.304 [0.235-0.353] |
| GigaVerbo-Text-Filter | 0.444 [0.385 - 0.489] | 0.324 [0.268-0.362] |
| Aroeira | 0.456 [0.401-0.499] | 0.346 [0.298-0.379] |
| FineWeb-2_Latn-por | 0.465 [0.421-0.503] | 0.348[0.318-0.373] |
| Aurora-PT (Our) | 0.465 [0.424-0.501] | 0.350 [0.324-0.374] |

Table 5: Descriptive statistics (Median and IQR) of the metrics Lexical Density and Stopword Ratio by corpus.

We calculated perplexity using the multilingual GPT model (Shliazhko et al., 2022) (Table 6). The results indicate that the corpus Aurora-PT achieved the lowest value compared to the other corpora analyzed. This performance suggests that the multilingual GPT model found it easier to predict the linguistic sequences of Aurora-PT, which may indicate greater internal coherence within the corpus.

| **Corpus** | **Perplexity** |
|---|---|
| brWaC | 107721.28 |
| GigaVerbo | 51199.35 |
| GigaVerbo-Text-Filter | 138990.48 |
| Aroeira | 306549.0 |
| FineWeb-2_Latn-por | 169300.78 |
| Aurora-PT (Our) | **20437.97** |

Table 6: Perplexity values for Brazilian Portuguese corpora.

However, it is important to emphasize that perplexity is not an entirely representative metric of corpus quality, as there is a possibility that the model was trained on documents belonging to the analyzed corpora, which would artificially reduce perplexity values.

### 4.2 NorBERTo's Performance Compared to Reference Models

In this section, we present an overview of NorBERTo's performance on Portuguese NLP tasks, comparing it with widely used reference models. Overall, NorBERTo achieved competitive results, standing out as state-of-the-art on the PLUE benchmark and delivering superior performance in inference tasks on ASSIN 2.

The benchmarks include ASSIN 2 for textual similarity and inference, and PLUE, a Portuguese adaptation of GLUE with tasks like RTE, WNLI, and MRPC. Additional benchmarks, such as FakeRecogna 2.0 and TweetSent-BR, are detailed in Appendix A. Most benchmarks are based on automatic translations from English, while those originally created in Portuguese are smaller, reflecting typical resource limitations.

Detailed results for ASSIN 2 and PLUE are presented in sections 4.2.1 and 4.2.2, including comparisons with models such as BERTimbau, XLM-RoBERTa, and Albertina.

#### 4.2.1 ASSIN 2

For the initial evaluation of NorBERTo, we addressed sentence similarity and textual entailment tasks. Sentence similarity was measured on a 0 to 5 scale using Pearson correlation, while entailment assessed whether one sentence logically follows from another (class 1 or 0), evaluated by F1-score. Both tasks used the ASSIN 2 benchmark (Real et al., 2020).

The ASSIN 2 dataset comprises approximately 10,000 sentence pairs, distributed into 6,500 instances for training, 500 for validation, and 2,448 for testing. This same split was used in the subsequent experiments.

For this evaluation, we carried out two experiments using four models derived from NorBERTo: the *sequence classification* and *cross encoder* variants, each trained on the *base* and *large* versions of NorBERTo. For comparison, we also included three reference models (*baselines*): BERTimbau (Souza et al., 2020), Albertina PT-BR (Santos et al., 2024), and XLM-RoBERTa (Conneau et al., 2019). All models were trained using the same hyperparameters, as detailed in Section 3.

The results for the entailment and similarity tasks are summarized in Table 7.

Overall, the results indicate that the NorBERTo sequence classification (large) variant achieved the best performance on the entailment task, with an F1-score of 90.4%, surpassing BERTimbau-large (88.1%). The NorBERTo cross encoder (large) variant showed similar performance (90.3%). On the other hand, for the similarity task, BERTimbau-large obtained the best result, with a correlation of 0.852, while the best configuration of NorBERTo reached 0.766.

Although the NorBERTo variants achieved competitive results on textual entailment, their performance on semantic similarity lagged behind the reference models. This difference is likely explained by the pretraining strategy: NorBERTo was trained entirely from scratch, whereas BERTimbau benefited from continued pretraining from English-initialized checkpoints. Prior work shows that initializing from large multilingual or English-pretrained encoders facilitates the transfer of syntactic and semantic knowledge learned from massive corpora (Devlin et al., 2018; Pires et al., 2019), often improving performance in fine-grained semantic tasks such as sentence similarity (Conneau et al., 2019; Liu et al., 2019).

#### 4.2.2 PLUE

For the second evaluation of NorBERTo, we used the PLUE corpus (Gomes, 2020), a translated version of the well-known GLUE benchmark (Wang et al., 2018). Three representative tasks were selected: MRPC, focused on paraphrase detection; RTE, for recognizing textual entailment; and WNLI, related to coreference and natural language inference. All tasks were evaluated using F1-score (macro, when applicable).

The experiment involved two variants of NorBERTo: *base* and *large*. For comparison, we also included two reference models (*baselines*): BERTimbau (Souza et al., 2020) and Albertina PT-BR (Santos et al., 2024). All models were trained using the same hyperparameters.

The results for the selected PLUE tasks are summarized in Table 8.

Overall, the results indicate that the NorBERTo-large variant achieved the best performance across all evaluated tasks. On MRPC,

| Model | Size | Entailment F1-score | Similarity Pearson Corr. |
|---|---|---|---|
| Albertina PT-BR | base | 0.874 | 0.826 |
| RoBERTa | base | 0.829 | 0.560 |
| BERTimbau | large | 0.8891 | **0.852** |
| BERTimbau | base | 0.8833 | 0.836 |
| mmbert | base | 0.8963 | 0.8212 |
| NorBERTo | base | 0.890 | 0.736 |
| NorBERTo (sequence classification) | large | **0.9038** | - |
| NorBERTo (cross encoder) | large | 0.9033 | 0.766 |

Table 7: General results for sentence inference and similarity tasks.

| Model | Size | MRPC | RTE | WNLI |
|---|---|---|---|---|
| BERTimbau | large | 0.8873 | 0.7545 | 0.5634 |
| Albertina PT-BR | base | 0.8779 | 0.6462 | 0.5493 |
| mmbert | base | 0.9051 | 0.7582 | 0.5633 |
| NorBERTo | base | 0.8926 | 0.7220 | 0.5774 |
| NorBERTo | large | **0.9191** | **0.7689** | **0.5774** |

Table 8: General results for PLUE tasks.

it reached 91.9%, surpassing BERTimbau-large (88.7%) by +3.2 points. On RTE, it achieved 76.9%, a gain of +1.4 points compared to BERTimbau-large (75.5%). For WNLI, NorBERTo-large obtained the highest score (57.7%), +1.4 points above the second-best model. The *base* version also delivered competitive performance, outperforming models of similar size and even larger ones, such as BERTimbau-large on MRPC.

These results confirm that NorBERTo establishes new state-of-the-art results on GLUE-like tasks in Portuguese.

## 5 Conclusion and Future Work

We presented Aurora-PT, a 331B-token corpus for Portuguese, and NorBERTo, a ModernBERT-style encoder trained from scratch on this corpus. Our experiments show that Aurora-PT is a large, lexically balanced and coherent Portuguese corpus. NorBERTo-large establishes new strong baselines on PLUE and achieves the best ASSIN 2 entailment F1 among the encoders we evaluated. Finally, NorBERTo is competitive on a variety of classification benchmarks as presented in Appendix A.

These findings reinforce that modern architectures combined with very large monolingual corpora can substantially improve encoders without requiring LLM-scale parameter counts. NorBERTo-base in particular demonstrates that a 150M-parameter model can surpass older 340M-parameter engines (BERTimbau-large) on several tasks.

However, no single model excels universally: BERTimbau-large and Albertina-900M still lead in semantic similarity, and XLM-R-large in some classification tasks. This diversity supports a toolbox approach, combining encoders, LLMs, and domain-specific models as needed.

Our work supports a right-sized AI perspective for Portuguese NLP: by combining high-quality, large-scale monolingual data with modern encoder architectures, we can obtain models that are powerful, efficient and easier to deploy and govern than general-purpose LLMs, while remaining complementary to them in hybrid systems.

Future work will include broader evaluations (e.g., QA, cross-lingual transfer, long-document tasks), scaling NorBERTo up and down with larger variants (1–2B parameters), improving similarity modeling with contrastive learning and using NorBERTo as retrieval backbone in retrieval-augmented generation (RAG) pipelines. We also plan to explore training other model architectures (decoder-only and encoder-decoder) with Aurora-PT. We will also release Aurora-PT and NorBERTo under open licenses.

## Limitations

Although the proposed model is based on the ModernBERT architecture and was trained exclusively on Portuguese-language documents, the total number of tokens used during pretraining—approximately 226 billion tokens, as computed using the original ModernBERT tokenizer—is considerably smaller than that employed in the training of the original ModernBERT models (Warner et al., 2024).

The ModernBERT authors report that both ModernBERT-base and ModernBERT-large were trained on the order of trillions of tokens, following initial warmup phases, resulting in a data scale substantially larger than that considered in this work (Warner et al., 2024).

This difference in corpus size may affect the model's generalization capacity and its performance on tasks that rely on broad factual knowledge or rarer linguistic patterns, as widely discussed in the literature on large-scale pretraining (Kaplan et al., 2020; Hoffmann et al., 2022). Accordingly, we emphasize that part of the observed performance or linguistic coverage limitations may stem directly from this reduced level of data exposure during the pretraining process.

## Acknowledgments

We thank the Instituto de Ciência e Tecnologia Itaú for the financial support that made this research possible. This study employed OpenAI's Deep Research to assist in the identification of relevant references, and OpenAI GPT and Microsoft Copilot to support the review and refinement of the manuscript.

## Conflict of Interest Statement

The opinions, findings, conclusions, and recommendations presented in this work are solely those of the authors and do not necessarily reflect the views of Itaú Unibanco or the Instituto de Ciência e Tecnologia Itaú. This document is not, and should not be interpreted as, investment advice or any form of investment service. It does not constitute, and should not be construed as, an offer to purchase or sell, a solicitation of an offer to purchase or sell, or a recommendation to purchase or sell any securities or other financial instruments. Furthermore, this research must not be used for commercial purposes. All data used in this study fully comply with the Brazilian General Data Protection Law (Lei Geral de Proteção de Dados – LGPD).

# A Additional Comparisons with the State-of-the-Art

As part of the comparison involving NorBERTo, we performed systematic hyperparameter optimization (HPO) for each model–dataset pair. Using Optuna's tree-structured parzen estimator (Akiba et al., 2019), we explored key hyperparameters such as learning rate and weight decay. Multiple runs were conducted for each configuration, and the best result, based on macro F1-score, was selected to represent each architecture. The optimal learning rate and weight decay values reported in Table 9 ensure reproducibility.

The experiments were based on four distinct datasets, each representing a specific NLP task in Portuguese: TweetSent-BR (Brum and Nunes, 2018): sentiment analysis in short messages; Hate-BR (Vargas et al., 2022): binary hate speech detection; TuPy-E (Oliveira et al., 2023): hate speech detection with classification by content type (multi-label); FakeRecogna 2.0 (Garcia et al., 2024): fake news detection.

As baselines, we included widely used monolingual and multilingual models, such as Albertina (Santos et al., 2024), BERTimbau (Souza et al., 2020), mmBERT (Marone et al., 2025), as well as the *base* and *large* variants of XLM-RoBERTa (Conneau et al., 2019). This diversity allowed us to assess the impact of model size and pretraining nature on final performance.

Table 9 presents the best results obtained by each model across the four datasets.

| **Model** | **Tweet-SentBR** | | | **Hate-BR** | | | **TuPy-E** | | | **FakeRecogna 2.0** | | |
|---|---|---|---|---|---|---|---|---|---|---|---|---|
| | F-score | LR | WD | F-score | LR | WD | F-score | LR | WD | F-score | LR | WD |
| NorBERTo-base | 0.7562 | 4.32e-05 | 0.0858 | 0.8986 | 1.01e-05 | 0.1007 | 0.8973 | 6.28e-06 | 0.1858 | 0.9841 | 2.10e-05 | 0.1871 |
| NorBERTo-large | 0.7607 | 6.45e-05 | 0.0929 | 0.9214 | 2.10e-05 | 0.2094 | 0.9053 | 2.27e-05 | 0.1796 | 0.9829 | 1.99e-05 | 0.0328 |
| Albertina-PTBR-100m | 0.7254 | 2.02e-05 | 0.1298 | 0.9071 | 4.79e-05 | 0.1266 | 0.8933 | 5.01e-06 | 0.0447 | 0.9800 | 2.29e-05 | 0.1298 |
| XLM-RoBERTa-base | 0.7517 | 2.32e-05 | 0.1521 | 0.9062 | 6.92e-05 | 0.0572 | 0.8950 | 1.04e-05 | 0.2513 | 0.9803 | 4.58e-05 | 0.2923 |
| XLM-RoBERTa-large | **0.7945** | 1.61e-05 | 0.0580 | **0.9300** | 2.46e-05 | 0.1440 | 0.9003 | 1.12e-05 | 0.0615 | **0.9843** | 1.63e-05 | 0.1964 |
| BERTimbau-base | 0.7552 | 5.87e-05 | 0.0451 | 0.9133 | 1.62e-05 | 0.2200 | 0.9055 | 1.08e-05 | 0.0222 | 0.9828 | 6.54e-05 | 0.1422 |
| BERTimbau-large | 0.7637 | 7.47e-06 | 0.0805 | 0.9271 | 6.52e-05 | 0.0821 | **0.9082** | 2.80e-05 | 0.1962 | 0.9838 | 9.55e-05 | 0.0905 |
| mmBERT-small | 0.7299 | 3.03e-05 | 0.2057 | 0.8862 | 1.36e-05 | 0.1913 | 0.8987 | 4.31e-05 | 0.0593 | 0.9820 | 4.47e-05 | 0.0304 |
| mmBERT-base | 0.7468 | 4.59e-05 | 0.1898 | 0.8986 | 4.59e-06 | 0.1407 | 0.9012 | 3.38e-05 | 0.2089 | 0.9823 | 4.08e-05 | 0.1003 |

Table 9: Best results on classification benchmarks.

Overall, the results show that larger models generally achieve higher performance. XLM-RoBERTa-large led on TweetSent-BR (79.5%), FakeRecogna 2.0 (98.4%), and Hate-BR (93.0%). On TuPy-E, the best performance was achieved by BERTimbau (90.8%). Nevertheless, NorBERTo-large remained highly competitive, closely matching the top models on Hate-BR (92.1% vs. 93.0%) and FakeRecogna 2.0 (98.3% vs. 98.4%), and outperformed several other monolingual and multilingual models. These findings reinforce that NorBERTo is robust and versatile, maintaining strong performance even against large-scale multilingual architectures.